\def\year{2019}\relax
\documentclass[letterpaper]{article} 
\usepackage{aaai19}  
\usepackage{times}  
\usepackage{helvet} 
\usepackage{courier}  
\usepackage[hyphens]{url}  
\usepackage{graphicx} 
\def\UrlFont{\rm}  
\usepackage{graphicx}  
\newcommand{\rpm}{\raisebox{.2ex}{$\scriptstyle\pm$}}
\title{Automated Let's Play Commentary }
\author{Shukan Shah$^1$, Matthew Guzdial$^2$ and Mark O. Riedl$^1$\\
$^1$School of Interactive Computing, Georgia Institute of Technology\\
$^2$Department of Computing Science, University of Alberta\\
shukanshah@gatech.edu, guzdial@ualberta.ca, riedl@cc.gatech.edu
}
 
\begin{document}

\maketitle

\begin{abstract}
Let's Plays of video games represent a relatively unexplored area for experimental AI in games.
In this short paper, we discuss an approach to generate automated commentary for Let's Play videos, drawing on convolutional deep neural networks. 
We focus on Let's Plays of the popular game Minecraft. 
We compare our approach and a prior approach and demonstrate the generation of automated, artificial commentary.
\end{abstract}
\section{Introduction}

Let's Plays have garnered an enormous audience on websites such as Twitch and YouTube. 
At their core, Let's Plays consist of individuals playing through a segment of a video game and engaging viewers with improvised commentary, often times not related to the game itself.
There are a number of reasons why Let's Plays may be of interest to Game AI researchers. 
First, part of Let's Play commentary focuses on explaining the game, which is relevant to game tutorial generation, gameplay commentary, and explainable AI in games broadly.
Second, Let's Plays focus on presenting engaging commentary. 
Thus if we can successfully create Let's Play commentary we may be able to extend such work to improve the engagement of NPC dialogue and system prompts. 
Finally, Let's Plays are important cultural artifacts, as they are the primary way many people engage with video games.

Up to this point Let's Plays have been drawn on for tasks like bug detection \cite{lin2019identifying} or learning game rules \cite{guzdial2017game}. 
To the best of our knowledge there have only been two attempts at this problem, the first focused on generation of a bag-of-words representation, which is an unordered collection of words that does not constitute legible commentary \cite{guzdial2018towards}.
The second attempt at this problem instead structured commentary generation as a sequence-to-sequence generation task \cite{li2019end}.
We do not compare against this second approach as it was not yet published during the development of this research.
In this paper we present an attempt at generating Let's Play commentary with deep neural networks, specifically a convolutional neural network (CNN) that takes in a current frame of a gameplay video and produces commentary. 
As an initial attempt at this problem we focus on Let's Plays of the game Minecraft.
We chose Minecraft due to its large and active Let's Play community and due to Minecraft's relative graphical simplicity.
In this paper we present two major contributions: (1) a dataset of Minecraft gameplay frames and their associated commentary and (2) the results of applying a CNN to this task, compared to the approach presented by Guzdial et al. \cite{guzdial2018towards}.
The remainder of this paper covers relevant prior work, presents our approach and implementation of the baseline, presents results of a brief quantitative analysis, and example output of our approach.

\section{Related Work}

This approach aims to take in video game footage (raw pixels) and output commentary.
To the best of our knowledge Guzdial et al. \shortcite{guzdial2018towards} were the first to attempt this problem.
Guzdial et al. focused on a preliminary approach towards Let's Play commentary of Super Mario Bros. gameplay, but notably could not produce full commentary.
Their approach focused on clustering pairs of Let's Play commentary utterances and gameplay video and then training non-deep machine learning models to predict a bag of words from an input gameplay frame.
More recently, Li et al. \shortcite{li2019end} represented artificial let's play commentary as a sequence-to-sequence generation problem, converting video clips to commentary.
Prior approaches have attempted to create commentary from logs of in-game actions for both traditional, physical sports games and video games \cite{kolekar2006event,graefe2016guide,barot2017bardic,Ehsan:2018:RNM:3278721.3278736,scores2017lee,Ehsan:2019:ARG:3301275.3302316}. 
These approaches depend on access to a game's engine or the existence of a publicly accessible logging system.

This work draws on convolutional neural networks (CNNs) to predict commentary for a particular frame of a gameplay video. CNNs have been employed to take an input snapshot of a game and predict player experience \cite{guzdial2016deep,liao2017deep}, game balance \cite{liapis2019fusing}, and the utility of particular game states \cite{stanescu2016evaluating}.

Significant prior work has explored Let's Play as cultural artifact and as a medium. 
For example, prior studies of the audience of Let's Plays \cite{sjoblom2017people}, content of Let's Plays \cite{sjoblom2017content}, and building communities around Let's Play \cite{hamilton2014streaming}. 
The work described in this paper is preliminary as a means of exploring the possibility for automated generation of Let's Play commentary. 
We anticipate future developments in this work to more closely engage with scholarship in these areas.

Other approaches employ Let's Play videos as input for alternative purposes beyond producing commentary.
Both Guzdial and Riedl \shortcite{guzdial2016game} and Summerville et al. \shortcite{summerville2016learning} employ Longplays, a variation of Let's Play generally without commentary, as part of a process to generate video game levels through procedural content generation via machine learning \cite{summerville2017procedural}. 
Other work has looked at eSport commentators in a similar manner, as a means of determining what approaches the commentators use that may apply to explainable AI systems \cite{dodge2018experts}.
Lin et al. \cite{lin2019identifying} draw on summarizing metrics of gameplay video including Let's Plays as a means of automatically detecting bugs, but do not directly engage with the video data.

\section{System Overview}

In this section, we give a high-level overview of two approaches (our approach and a baseline) for automated commentary generation. 
We describe our implementation of the baseline approach as originally discussed in Guzdial et al. \cite{guzdial2018towards}. 
The baseline can be understood as a variation of our approach in which we test the assumption that training on clustered subsets of data reduces the variance of Let's Play commentary and consequently improves commentary prediction. 
We first delve into the preprocessing steps to extract and featurize our data for the experiments. 
We then describe the two approaches in succession.

In an idealized final version of this approach, first Let's Play videos would be collected with their associated commentary.
Second, this data would be preprocessed to featurize the data.
Third, this data would be used to train a convolutional neural network model.
Finally, this new model would be fed in new video and produce output commentary.

\subsection{Dataset}

For our dataset, we collected three 25-minute YouTube videos, one each from three popular Minecraft Let's Players. 
We extracted the associated text transcripts for each of these videos generated by YouTube to serve as our commentary corpus. 
We applied ffmpeg, an open source tool for processing multimedia files, to each video to break apart each video into individual frames at 1 FPS. 
Although we observed that each sentence in the commentary usually spanned a few frames, we purposely paired each frame with a sentence. 
In other words, there were multiple frames-comment pairs with the same commentary. 
We did this for simplicity's sake so that it would be easier for our model to learn the relationship between single frame-comment pairs. 
We refrained from converting the images to grayscale to prevent the loss of any color features. 
This is especially important for a game like Minecraft, in which all game entities are composed of cubes that primarily differ according to color. 
In total, our dataset is comprised of 4840 frame-comment instances, 3600 of which were used for our training set and the rest for our test set.\footnote{This dataset is publicly available at: https://github.com/shukieshah/AutoCommentateDataset.}

\begin{figure*}[tb]
\centering
	\includegraphics[width=5in]{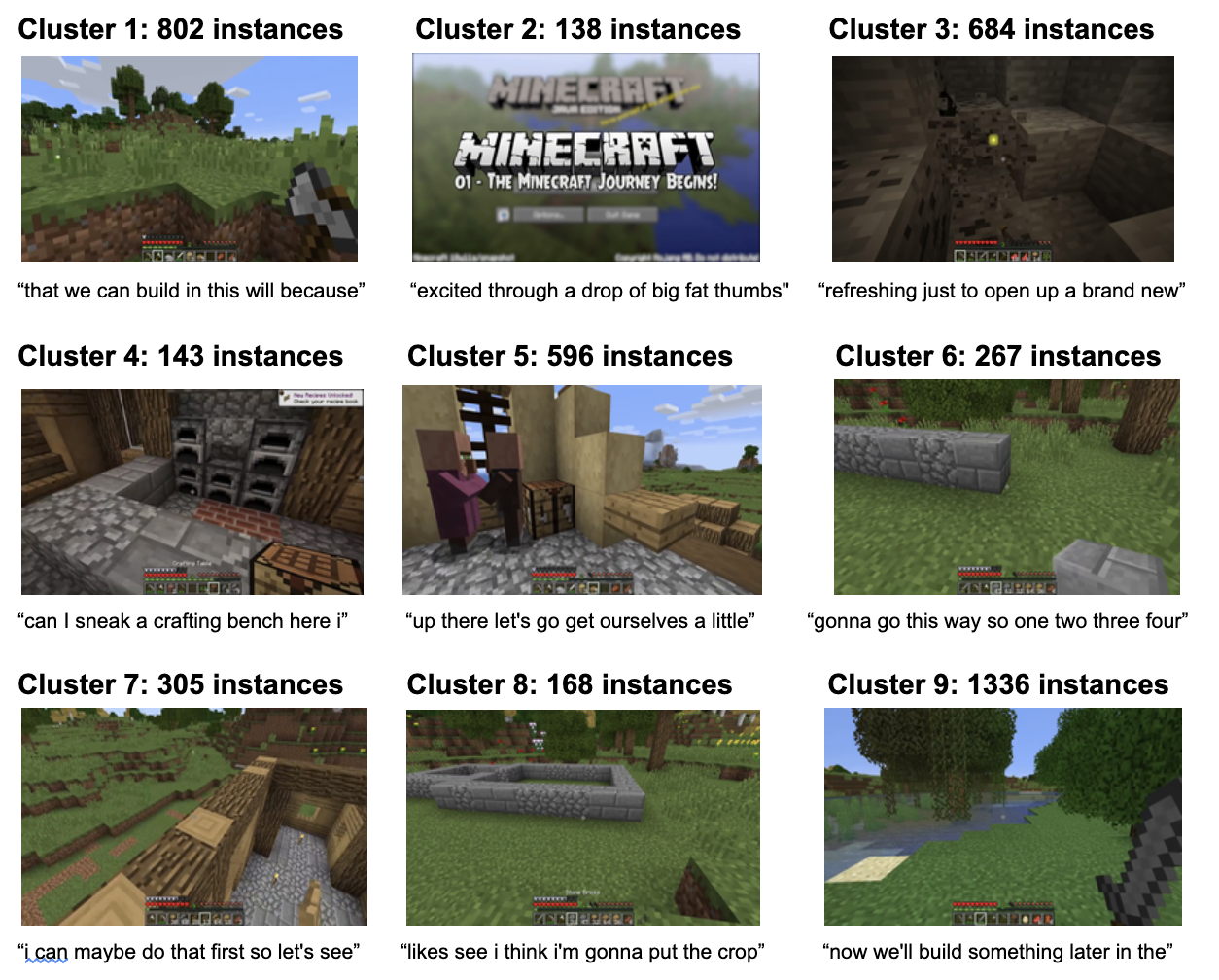}
	\caption{The medoids of each of the clusters found by the K-Medoids clustering algorithm.}
	\label{fig:clusters}
\end{figure*}

\subsection{Sentence Embeddings}

Sentence embeddings are a standard way in the field of natural-language processing (NLP) to represent sentences in a vector-representation appropriate to deep neural networks.
We tokenized the sentence in each frame-comment pair and converted it to a 512-dimensional numerical vector using the Universal Sentence Encoder \cite{cer2018universal}. 
The Universal Sentence Encoder is a model that is trained with a deep averaging network (DAN) encoder to convert plain English strings into a corresponding vector representation. 
We used this representation over traditional Word2Vec word embeddings because the model is specifically catered towards 'greater-than-word' length strings such as the sentences and phrases present in our dataset. 
The sentence embeddings produced through this method are better able to model contextual awareness in sequences of words, which is crucial for the use case of commentary generation.

\subsection{Our Approach}

For our approach, we trained a convolutional neural network (CNN) with the 4840 training instances, taking as input the gameplay frame and predicting the associated commentary in a sentence embedding representation.
The CNN architecture was as follows: (1) a conv layer with 32 3x3 filters followed by a max pool layer, (2) a second conv layer with 64 3x3 filters, (3) a third conv layer with 64 3x3 filters followed by a max pool layer, (4) a fully connected layer of length 1024, (5) a dropout layer fixed at 0.9, (6) a fully connected layer of length 512, which represents the final 512-vector sentence embedding. 
We used adam \cite{kingma2014adam} for optimization (with a learning rate of 0.001) and mean-square error for our loss function. 
All layers used leaky ReLU activation \cite{xu2015empirical}.
We employed Tensorflow \cite{abadi2016tensorflow} and trained until convergence on our training set (roughly 20 epochs). 
We note that this architecture was constructed by considering architectures for similarly sized datatsets for image captioning (including CifarNet \cite{hosang2015taking}), a related area for computer vision given that the Let's Play utterance can be thought of as like an abstract caption for the gameplay frame.

\section{Baseline}

The baseline, adapted from Guzdial et al. \cite{guzdial2018towards}, is ironically more complex than our approach.
This approach calls for first clustering the Let's Play data as frame and utterance pairs and then training a unique machine learning model for each cluster individually. 
Thus we first cluster our 4840 training instances and then train the same CNN architecture used on our approach on the largest of the output clusters. 
We walk through this process in greater depth below.

\subsection{Image Embeddings}

For the clustering of the frame and utterance data we re-represent our gameplay frames in an image embedding representation.
Image feature embeddings are similar to the sentence embeddings discussed above. 
These vectors were generated by passing images through a ResNet \cite{targ2016resnet} CNN architecture trained on the ImageNet dataset \cite{deng2009imagenet}. 
The images were fed through the network up to the penultimate activation layer, and the activation weights were extracted as features for the particular image.
This allowed us to better represent the context of the image for clustering purposes without having to directly compare images to one another, which would have been highly time consuming.

\subsection{Clustering}

Using the process described in \cite{guzdial2018towards} we employed K-medoids clustering with $K$ estimated via the distortion ratio, using means square error as the distance function for the clusters, comparing both image and sentence vectors combined into a single vector.

Figure \ref{fig:clusters} shows the actual medoid instance (frame-comment pairs) for each of the learned clusters.
It is interesting to note that the the clusters with the most instances (cluster 9 and 1 respectively) comment on 'building' things, a key component of Minecraft. 
Furthermore, the clustering seems to have chosen clusters that capture unique moments of gameplay. 
For example, cluster 2 represents an opening-screen where Let's Players typically introduce themselves and greet the audience. 
Cluster 3, on the other hand, represent underground gameplay which is distinctive both visually and mechanically. 
From a qualitative standpoint, the clusters appear to effectively capture high-level themes.
Thus we find it to be a successful implementation of the Guzdial et al. work \cite{guzdial2018towards}.

\begin{figure*}[tb]
\centering
	\includegraphics[width=6in]{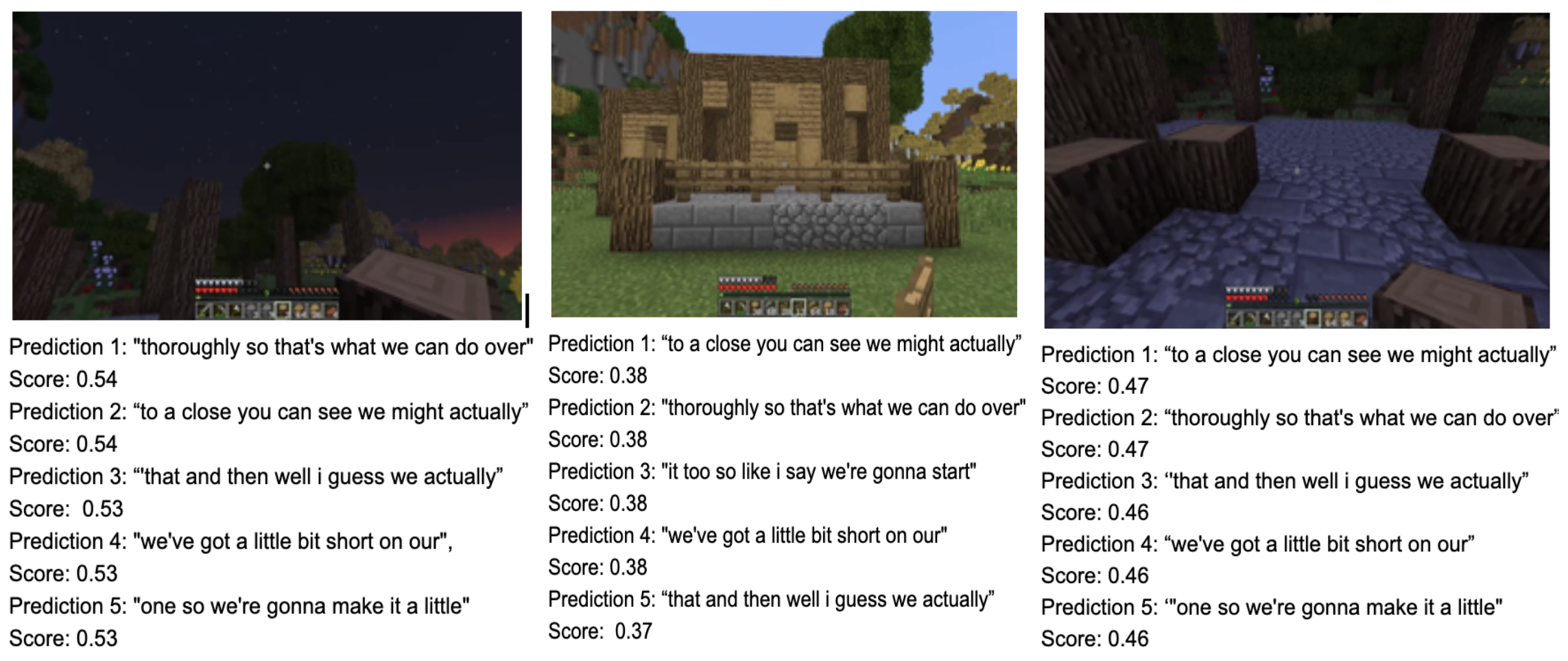}
	\caption{Each frame is paired with the five closest nearest-neighbors of the model's actual predicted commentary.}
	\label{fig:examples}
\end{figure*}

\section{Evaluation}

\begin{table}[tb]
\begin{center}
\caption{Average percentile error of our approach and the three largest clusters for the baseline. }
\begin{tabular}{|l|c|c|} 
 \hline
 Model & Percent Error & Training Set Size\\
 \hline
 CNN & \textbf{0.961\rpm0.026} & 4840\\
 \hline
 Cluster 9 CNN & 0.977\rpm0.023 & 1336\\
 \hline
 Cluster 1 CNN & 0.975\rpm0.042 & 802\\
 \hline
 Cluster 3 CNN & 0.980\rpm0.024 & 684\\
 \hline
\end{tabular}
\end{center}
\label{tab:results}
\end{table}

Table 1 compares the results of our approach and the baseline approach for the three largest clusters in terms of the average percent error on the test set. 
By average percent error we indicate the averaged percentile error for each predicted utterance compared to the true utterance across the test set.
Thus lower is better. 
The lowest possible value of this measure would then be 0.0 and the highest (and worst) value would be 1.0.
As one can see in Table 1, none of the approaches do particularly well at this task. This underscores the difficulty in predicting natural language labels given gameplay frame video only. 
However, we note that our approach outperforms the baseline across all three of its largest clusters.
All of the other baseline per-cluster approaches do strictly worse and so we omit them from our analysis.

\section{Example Output}

Figure \ref{fig:examples} shows the predicted commentary and cosine similarity scores for three test images for our approach. 
We include the closest sentences from our training set to the predicted sentence encoding as novel commentary due to the limitations of the Universal Sentence Encoder \cite{cer2018universal}, but with another sentence embedding we could directly output novel commentary.
The commentary represents the five closest neighbors to the actual predicted output from the baseline model. 
As one can see, there are repeats of predicted sentences across instances. 
This is because we are only retrieving commentary from within our training dataset which may bias certain sentences due to their greater overall semantic similarity to other sentences. 
The ordering and scores of the predictions vary for different test instances, indicating that the model did not just learn a single strategy.
Although the commentary doesn't correlate well to the images shown, the generation of commentary is a promising advancement from prior work.

\section{Conclusions and Future Work}

In this paper we demonstrate an initial approach to Let's Play commentary for the game Minecraft.
While the initial results are not particularly impressive, they outperform the original approach to this problem.
We did not compare to the more recent Li et al. \shortcite{li2019end} as it was unavailable during our research, but we note that they represent the problem in a significantly different way, making a direct comparison non-trivial.
Further, the results speak to the difficulty of this problem, which we anticipate being a fruitful area of future research.
Our primary contributions are our dataset of Minecraft Let's Play frames and associated commentary and the results and analysis presented in this paper. 

This work had a number of limitations, which we hope to address in future work.
First, we acknowledge a limitation in the relative weakness of the results.
We imagine two major reasons for this issue: (1) that the model makes predictions without knowledge of previous utterances and (2) the size of the training dataset. 
Thus we anticipate greater success by including as input to the model the prior utterance as a sentence embedding and increasing the size of the training dataset.
The second limitation we identify is in our choice to limit our output to a single game. 
While we acknowledge that this is helpful for an initial approach, an ideal system could take in any arbitrary gameplay video. 
Further, increasing the games we include would help us solve the training dataset size problem. 
Nonetheless, generalizing to other types of games would itself present a unique challenge since context and commentary are highly dependent on the rules and design of a particular game.  
Although solving this problem is nontrivial, in future work we hope to extend this project to other, popular games for Let's Plays by abstracting lower level details and focusing on higher level themes shared across games.

\section{Acknowledgements}

This material is based upon work supported by the National Science Foundation under Grant No. IIS-1525967. Any opinions, findings, and conclusions or recommendations expressed in this material are those of the author(s) and do not necessarily reflect the views of the National Science Foundation.

\bibliographystyle{aaai}
\bibliography{aaai}
\end{document}